# Image Segmentation using Unsupervised Watershed Algorithm with an Over-segmentation Reduction Technique

Ravimal Bandara[1]

[1] Faculty of Information Technology, University of Moratuwa, Sri Lanka

E-mail: ravimalb@uom.lk

**Abstract**

Image segmentation is the process of partitioning an image into meaningful segments. The meaning of the segments is subjective due to the definition of homogeneity is varied based on the users' perspective hence the automation of the segmentation is challenging. Watershed is a popular segmentation technique which assumes topographic map in an image, with the brightness of each pixel representing its height, and finds the lines that run along the tops of ridges. The results from the algorithm typically suffer from over segmentation due to the lack of knowledge of the objects being classified. This paper presents an approach to reduce the over segmentation of watershed algorithm by assuming that the different adjacent segments of an object have similar colour distribution. The approach demonstrates an improvement over conventional watershed algorithm.

**1. Introduction**

Image segmentation is the process of partitioning an image to meaningful segments. There are many segmentation algorithms available, but nothing works perfect in all the cases. In computer vision, Image segmentation algorithms available either as interactive or automated approaches. In medical imagine, interactive segmentation techniques are mostly used due to the high precision requirement of medical applications. But some applications like semantic indexing of images may require fully automated segmentation method.

Automated segmentation methods can be divided into two categories namely supervised and unsupervised segmentation[1], [2]. In supervised segmentation the method uses a machine learning technique to embed the knowledge of objects (or ground truth segments) in images to the segmentation algorithm. This method mimics the procedure which a human follows when he hand segment an image. It is a difficult task to give a robust and powerful knowledge and experience to a machine. The current technologies enables creating supervised segmentation only for a specific domain of images[3]. Hence unsupervised segmentation methods are widely used in general applications.

Unsupervised segmentation may use basic image processing techniques to complex optimization algorithms. Hence these segmentation methods take much more time when a highly accurate results are in need. The main problem in unsupervised segmentation algorithms is the difficulty of balancing the over-segmentation and under-segmentation.

This article explains an implementation of unsupervised watershed algorithm for image segmentation with a Hue-Saturation histogram matching technique to reduce the over-segmentation of the segmentation algorithm. The rest of the article is arranged in the following manner. First, the background of this approach incluing the related works is discussed and then the details of aprroah is presented. The results are then presented with the discusssion followed by the conclusion.

## 2. Background

The algorithm includes a few concepts namely, Otsu thresholding, morphological opening, distance transformation, watershed algorithm, hue-saturation histogram and histogram comparison. The following subsections describe the each of the concepts briefly.

*2.1 Watershed Segmentation Algorithm*

Watershed segmentation is a nature inspired algorithm which mimics a phenomenon of water flowing through topographic relief. In watershed segmentation an image is considered as topographic relief, where the gradient magnitude is interpreted as elevation information. Watershed algorithm has been improved with marker-controlled flooding technique [4]. In this implementation, the selection of the markers has been automated by extracting brighter blobs in the image. The brighter blobs can be extracted by refining the binarized version of the image using a suitable threshold. This implementation uses an automatic threshold selection algorithm namely Otsu in order to obtain the binary image out of the grayscale image.

*2.2 Otsu Thresholding*

Image thresholding is one of the methods to convert an gray-scale image to a binary. Selecting a good threshold value is the main problem in any thresholding technique[5]. Otsu thresholding is one of the automatic thresholding methods available.

Otsu's threshold iterate through all the possible threshold values to find the threshold value where the sum of foreground and background spreads is at its minimum. The measure of spreads involves in several simple calculations which have been explained in [6].

The binarized images typically contain many small foregrounds which may yeild large number of segments. The smaller foregrounds can be removed by applying morphological operations.

*2.3 Morphological Opening*

Morphological opening is an application of some primary operators in image processing namely morphological erosion and dilation. The effect of an opening is more like erosion as it removes some of the bright pixels from the edges of regions and some small isolated foreground (bright) pixels. The difference between the basic erosion and opening is that the morphological opening is less destructive than the erosion operation. The reason is that the opening operation is defined as an erosion followed by a dilation using the same structuring element (kernel) for both operations. To read more about morphological opening, refer [7].

Sometimes some regions are more meaningful if it is partitioned to several subregions. In the case of splitting the regions that has been formed with several subregions connected through a small line or a bunch of pixels, the distance transformation can be used.

*2.4 Distance Transformation*

Distance transformation is an operation which typically applies on binary images to detach two or more convex forgrounds. The transform results in a graylevel image that is similar to the input image in shape, but the graylevel intensities of points inside foreground regions are varied to show the distance to the closest boundary from each point. [8] explains the details of implmentation of the distance transformation.

The distance transformation provides some early version of markers to initiate the segmentation algorithm. However, the segments may have same color distribution, in which case, we can assume the regions belong to the same object. The color distribution can be obtained by using a color histogram.

*2.5 Hue-Saturation Histogram*

In computer vision, there are numerous methods available to measure the visual similarity of two regions in an image. The matching technique is actually should selected based on the application. For some applications which has objects with unique textures can use a good texture descriptor. If the set of objects has unique shape then a shape descriptor can be used. In general cases a set of well known descriptors such as SIFT[9] or SURF[10] can be used. But in most cases using a sophisticated description for measure the similarity is not practical due to high computational complexity. Most of the images in general applications (like semantic indexing of images in web) have objects with objects in many kinds of textures, shapes and colors. If we can assume that there is a great probability to have the same spread of colors in different regions in a single object then we can use the color feature to measure the similarity of different segments of an object.

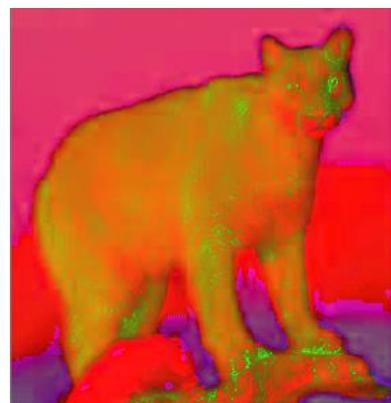

*Figure 1 A sample Image in HSV Color Space*

One of the best ways to express the color spread is using a histogram. But histograms of images in RGB color space is not always represent the color spread well enough to all applications. Histograms in HSV (Hue, Saturation and Value) color space is known to perform well in all most all applications that the color information is dominant. The hue component of a color corresponds to a one major color where as Saturation and Value corresponds to variations of the colors in Hue component. Figure 1 clearly shows that the color spreads in different regions (we extracted using the watershed segmentation) in the same object are similar. Therefore in this implementation a Hue-Saturation histogram is extracted for each region and measure the similarity using a measurement called Bhattacharyya distance.

*2.6 Bhattacharyya Distance*

In statistics, the Bhattacharyya distance quantify the similarity of two probability distributions. It can be used to determine the relative closeness of the two samples being considered. It is considered to be more reliable than several other measurements such as Mahalanobis distance. Details of calculating Bhattacharyya distance can be found in [11]. The next section describes the proposed approach in detail.

## 3. Proposed Method

The goal of the proposed method is to improve the watershed algorithm by providing an unsupervised over-segment reduction technique. However, the method utilizes an existing technique for seed the algorithm with automatically obtained markers.

*3.1 Automatic Watershed Seed Generation*

To generate the automatic markers for seedinf the watershed algorithm, we first convert the color image to grayscale. Then we apply the following steps:
1. Binarize the grayscal image by using Otsu thresholding.
2. Remove smaller foregrounds by using morphological opening.
3. Detach any attached convex foregounds by applying the distance transformation.
4. The resultant image is again thresholded for completing the detachment.
5. Feed the markers to the watershed algorithm

Once the watershed segmentation applied, it provides as many segmentation as the number of markers. The next step is to reduce these segments by identifying the oversegments.

*3.1 Similarty Matching of Adjacent Segments*

The core assumption of this work is that the adjacent segments that belongs to the same object should have nearly similar colour distribution. There Hue-Saturation histogram is calculated for each of the segments through the following steps.
1. Obtain mask for different segments.
2. Create a copy of the original image and convert it to HSV colour space.
3. Drop the value channel from the image.
4. Calculate Hue-Saturation histogram for each region by using the segment mask.
5. The calculate the Bhattacharyya distance between each of the adjacent histograms.
6. If the distance between two adjacent segments is less than a threshold, the pair will be merged.

Once the adjacent segments with similar histograms are merged, the result can be visually verified. The result of each of the steps is presented in the next section.

## 4. Results and Discussion

This study used several random images downloaded from google image for measure the performance of the method qualitatively. The different stages and the final result of the segmentation will be presented using a one random image which was used in the study.

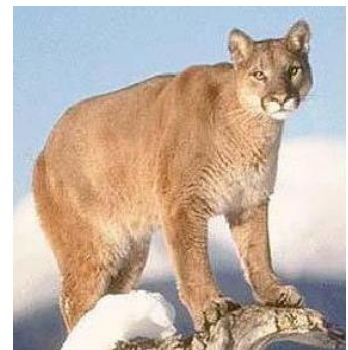

*Figure 2 Original Image*

Figure 2 shows the original image in which a Cougar Kitten poses in front of a snowy background. The image is then fed to the algorithm without providing any details about the background and foreground.

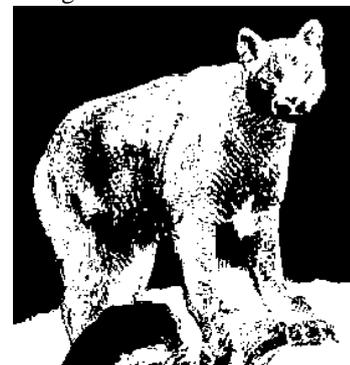

*Figure 3 Result after thresholding with Otsu Threshold*

Figure 3 Shows the binary image in which the foreground is shown by using white pixels. However not the whole foreground is having white pixels and also some background area also has white pixels.

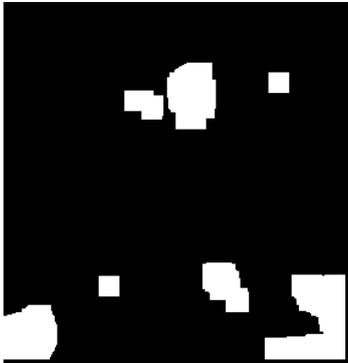

*Figure 4 After Applying Morphology-Opening to the Otsu-Thresholded Image*

Figure 4 Shows the initial points such that the seeds of the segments. In this example there are no any attached convex foregrounds. However, the algorithm is executed blindly, hence this input is going through the distance transformation and re-thresholding.

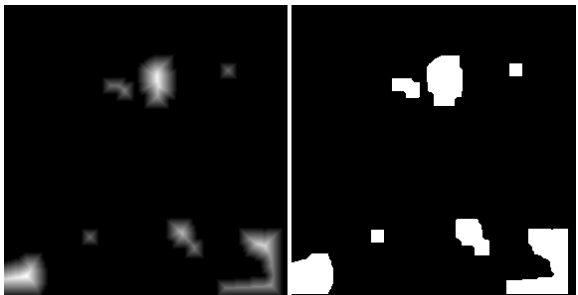

*Figure 5 The distance transformed image (left side) and its thresholded image*

Figure 5 shows the result of distance transformation and its thresholded binary image. As there were no any attached convex foregrounds, the resultant image is having the same set of foregrounds appeared in Figure 4.

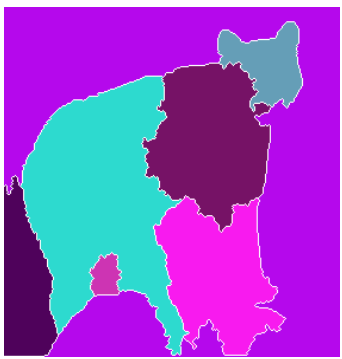

*Figure 6 Result of watershed algorithm initialized by the generated seeds.*

Figure 6 clearly shows that the foreground is highly oversegmented where the background also has slight over-segmentation. Note that the colors in different segments are random and only for identifying different segments.

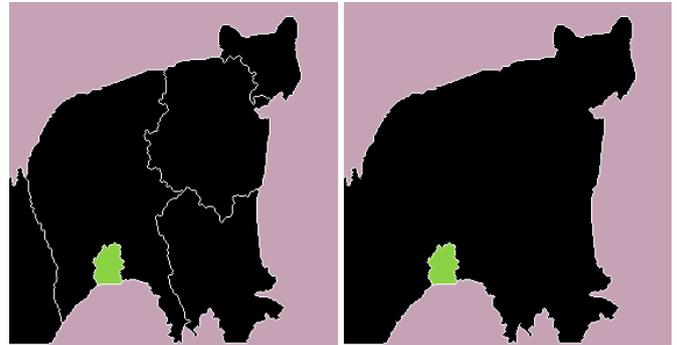

*Figure 7 Resultant image after merging the segments having similar histograms (**Left**: with the boundaries of merged segments; **Right**: with the boundaries of refined segments)*

Figure 7 clearly shows that there are only three segments in this image where two of them accurately covers the background (in pink pixels) and the foreground (in black pixels). When comparing the similarity of Hue-Saturation histogram of different regions, the distance threshold was set to 0.8 where it gave a reasonable performance over majority of the selected images.

## 5. Conclusion

This article presents an improvement for watershed algorithm by reducing the over segmentation by considering the color distribution of each of the segments. The results demonstrate that although the algorithm take only the image as its input, the resultant segments shows a great relationship between the refined segments and the semantically meaningful objects in the image compared to the raw output of the watershed algorithm. The future enhancement may be directed to evaluate more features including the texture for detect over segmentation. The source-code is available at https://www.codeproject.com/Articles/751744/Image-Segmentation-using-Unsupervised-Watershed-Al.